\documentclass{article}
\usepackage{ijcai22}
\UseRawInputEncoding
\usepackage{times}
\usepackage{soul}
\usepackage{url}
\usepackage[hidelinks]{hyperref}
\usepackage[utf8]{inputenc}
\usepackage[small]{caption}
\usepackage{graphicx}
\usepackage{amsmath}
\usepackage{amsthm}
\usepackage{amsfonts}
\usepackage{booktabs}
\usepackage{algorithm}
\usepackage{algorithmicx,algpseudocode}
\newcommand\Algphase[1]{%
\vspace*{-.7\baselineskip}\Statex\hspace*{\dimexpr-\algorithmicindent-2pt\relax}\rule{\columnwidth}{0.4pt}%
\Statex\hspace*{-\algorithmicindent}\textbf{#1}%
\vspace*{-.7\baselineskip}\Statex\hspace*{\dimexpr-\algorithmicindent-2pt\relax}\rule{\columnwidth}{0.4pt}%
}
\usepackage{subcaption}
\usepackage{multirow}
\usepackage{multicol,tabularx}
\title{Physics Constrained Flow Neural Network for Millisecond-timescale Predictions in Data Communication Networks}

\author{
Xiangle Cheng$^1$
\and
Shihan Xiao$^1$\and
Yingxue Zhang$^{2}$\and
Zhitang Chen$^3$\And
Pascal Poupart$^4$
\affiliations
$^1$Huawei 2012 Network Technology Lab\\
$^2$Huawei Technologies Canada\\
$^3$Huawei Noah’s Ark Lab\\
$^4$University of Waterloo
\emails
\{chengxiangle1, xiaoshihan, yingxue.zhang, chenzhitang2\}@huawei.com,
ppoupart@uwaterloo.ca
}

\begin{document}

\maketitle

\begin{abstract}
  Machine learning is gaining momentum in various recent models for the dynamic analysis of information flows in data communication networks. These preliminary models often rely on off-the-shelf learning techniques to make predictions based on historical statistics while disregarding the physics governing the generative process of these flows. Instead, this paper introduces \textbf{Flow} \textbf{N}eural \textbf{N}etwork (FlowNN) to improve short-timescale predictions with physical knowledge. This is implemented by embedding the logics of network congestion control and hop-by-hop forwarding into the data encoding layer. A self-supervised learning strategy with stop-gradient is also designed to improve the transferability of the learned physical logics. For milisecond-timescale flow prediction tasks, FlowNN decreases the loss by 17\% $\sim$ 71\% in comparison to state-of-the-art baselines on both synthetic and real-world networking datasets, which shows the strength of this new approach. Code will be made available.
\end{abstract}

\section{Introduction}\label{Introduction}
\begin{figure}[t]
\centering
\includegraphics[width=.99\columnwidth]{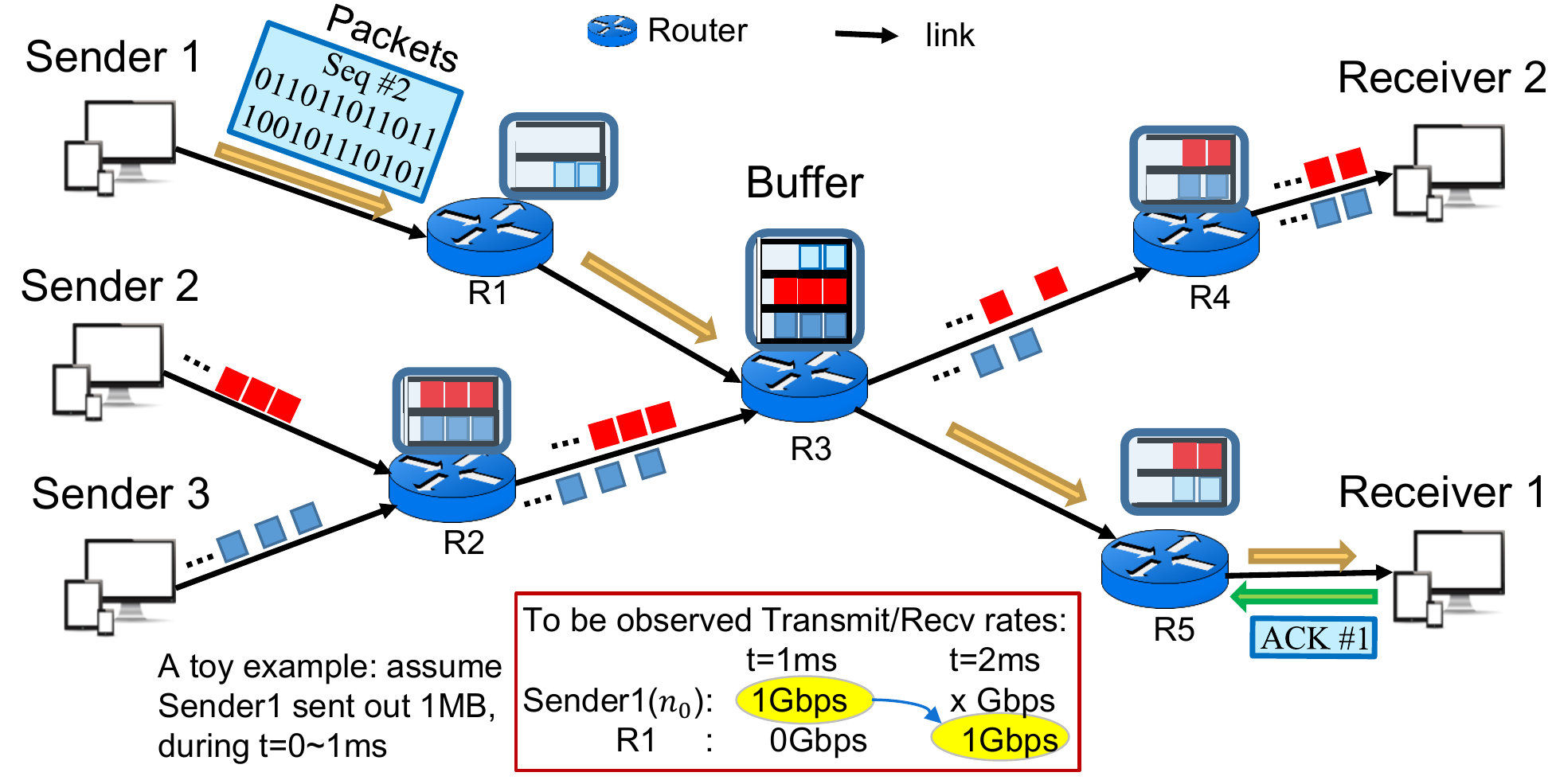} 
\caption{Packet flows in data communication networks. It is better to predict the bottom-right $1~Gbps$ by referring to only the upper-left $1~Gbps$ and excluding the information of $0~Gbps$ at the bottom left.}
\label{Packet_flows_in_IP_networks}
\end{figure}

Data communication networks provide the majority of data transmission services to support today's internet applications and present huge social value. This paper proposes a dynamic analysis model to deal with flows of information from sources to destinations in data communication networks. As depicted in Fig.~\ref{Packet_flows_in_IP_networks}, the source of a flow is a node (computer, phone, router/switch, etc.) in the network from which packets\footnote{Information is encapsulated in packets, which can be seen as particles travelling in the network.} start their travel and the destination is where they end. Throughout the lifespan of a flow (e.g., a 10-minute phone call or a 2-hour online video), packets consistently travel along the assigned routing path connecting the source and destination. Depending on the real-time congestion conditions of nodes on the routing path, packets may experience different buffering or retransmission delays, resulting in varying transmission rates and service delays.

A good understanding of the behavioral patterns of the underlying packets and flows plays a crucial role in network planning, traffic management, as well as optimizing Quality of Service (QoS, e.g., transmission rate, delay, etc.). However, the highly nonlinear dynamics of the networking system often create challenges for a fine-grained analysis. Particularly, at the milisecond timescale, the traffic traces are highly skewed and it is difficult to recognize obvious patterns.

Therefore, recent works often rely on off-the-shelf machine learning models to make predictions based on the data statistics of a flow. The remarkable advances achieved in long-timescale prediction tasks (e.g., at seconds or minutes intervals \cite{Kalander2020,CNNdatacenter2018}) are a testament to this minimalist principle. A key limitation of these machine learning models is that they disregard the physics governing the generative process of these packet flows.

In practice, the dynamics of spatio-temporal flows are governed by physical laws or models regulating the networking system. For instance, in networks with a TCP/IP protocol suite, the real-time transmission rates of different packet sources are typically modulated dynamically by congestion control models to prevent congestive collapse \cite{Nathan_Congestion_Control2019}. The conservation law describes the macroscopic fluid dynamics of data flows during hop-by-hop transmissions \cite{flowPhysicModel2008}. The physics establishes an \textit{inductive bias} \cite{InductiveBias2018} when predicting from history. That is, the future is not entirely random or \textit{noisy} even when observed at a millisecond granularity, but can be predicted from physical models or laws.

As illustrated in Fig.~\ref{Packet_flows_in_IP_networks}, assume $1MB$ of data is sent out from the source node $n_0$ (i.e., Sender1) at the first time interval $t=1ms$. After a certain delay (e.g., $1ms$) due to link propagation and packet processing, the first $1MB$ of data will arrive at its next
routing node $R_1$ at $t=2ms$. In this case, at $t=1ms$, the sending rate at the source node $n_0$ and the receiving rate at router $R_1$ are $x_{r,n_0}^{t_1}=1Gbps$ and $x_{r,R_1}^{t_1}=0Gbps$, respectively. At $t=2ms$, the associated rate at $R_1$ becomes $x_{r,R_1}^{t_2}=1Gbps$, while $x_{r,n_0}^{t_2}$ is again determined by the congestion control model according to the acknowledged ACK information. Such forwarding process will continue hop by hop until reaching the destination (Receiver1).

The above hop-by-hop data transmission establishes a special connection between space and time. We can observe in Fig.~\ref{Packet_flows_in_IP_networks} that $x_{r,R_1}^{t_2}=1Gbps$ is \textbf{\textit{physically}} determined by $x_{r,n_0}^{t_1}=1Gbps$, the history information of its predecessor node (i.e, source $n_0$) rather than its own history record $x_{r,R_1}^{t_1}=0Gbps$. Therefore, such time-resolved flow data not only tells us \textit{who} is related to whom, but also \textit{when} and in which \textit{order} relations occur. This provides a specific relational bias for modelling the spatio-temporal evolution of such flow data.

In this paper, we investigate how to embed such essential physical bias in a learning model to improve the short-timescale (millisecond-level) flow predictions. Accordingly, we propose \textbf{Flow N}eural \textbf{N}etwork (FlowNN), the first customized learning model for the data analysis of networking flows, while respecting packet routing structures and network mechanisms.

Our contributions can be summarized as follows: (1) We develop FlowNN, a prediction model that exploits data correlations physically induced by the congestion control model and flow conservation rule during hop-by-hop data forwarding; (2) We design a self-supervised learning strategy with stop-gradient that can learn the feature representation and physical bias simultaneously at each gradient update step; (3) We show with realistic packet datasets that FlowNN achieves consistently better results and outperforms the incumbent baselines by reducing the loss by 17\% $\sim$ 71\% for short-timescale prediction problems.

\section{Preliminaries}
\begin{figure}[t]
\centering
\begin{subfigure}[b]{1\columnwidth}
\includegraphics[width=.99\columnwidth]{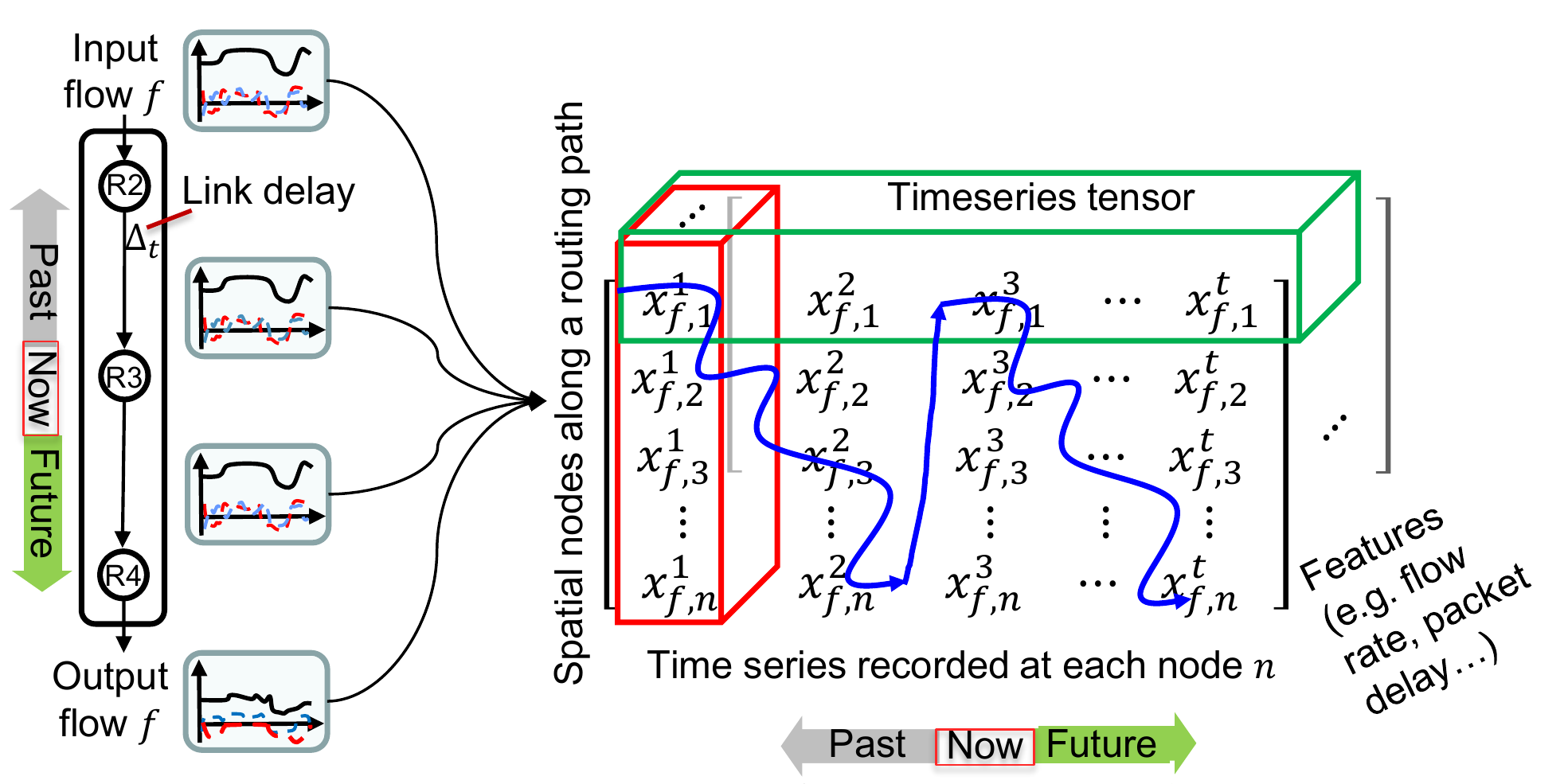} 
\caption{}
\label{fig_flowtensor_a}
\end{subfigure}
\centering
\begin{subfigure}[b]{0.99\columnwidth}
\includegraphics[width=.99\columnwidth]{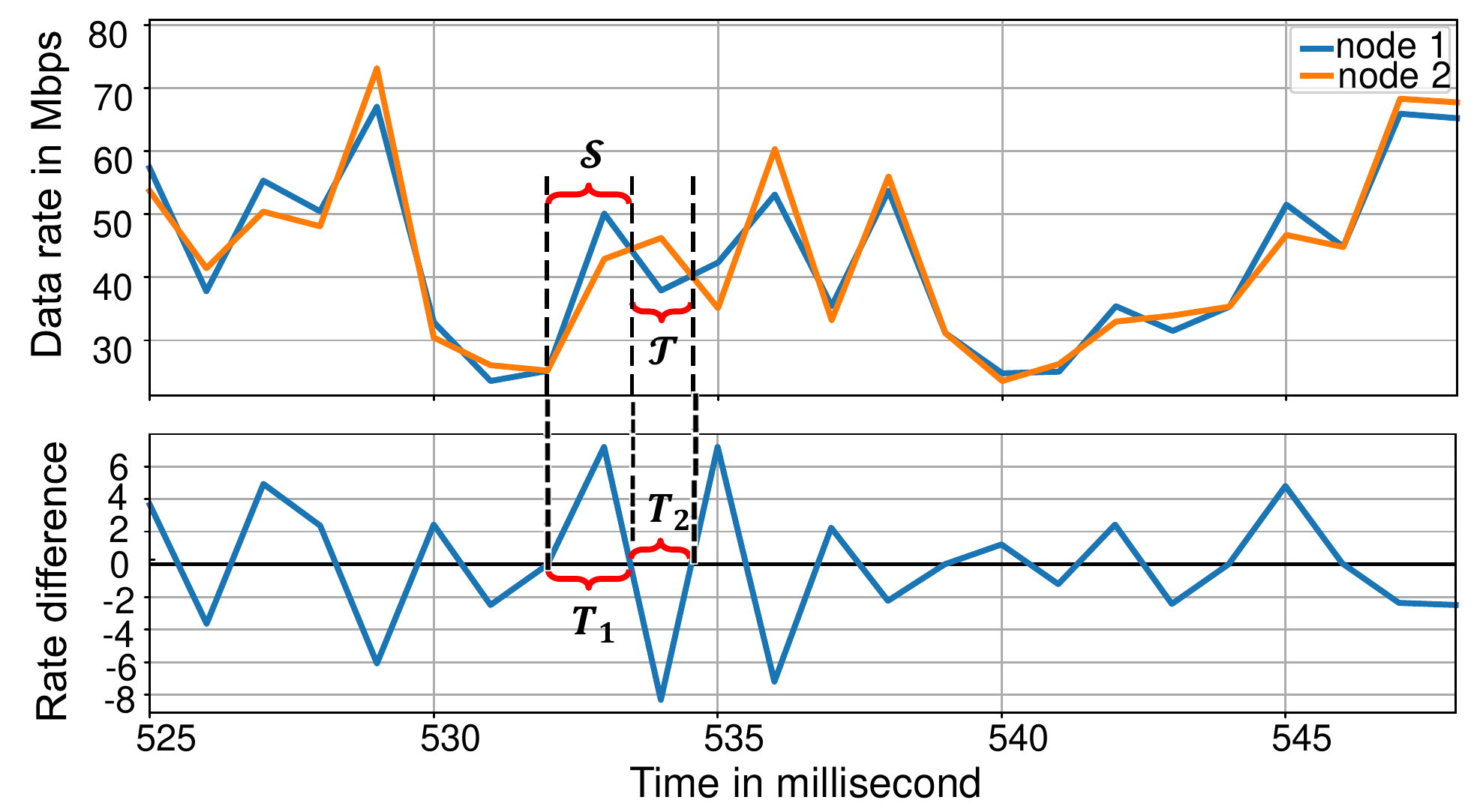} 
\caption{}
\label{fig_flowtensor_b}
\end{subfigure}
\caption{(a) Data of a packet flow in the form of time-series tensor. The blue curve characterizes the data correlations propagated along the space-time dimensions; (b) A sample of data rates of a flow and the observed rate difference between two neighboring routing nodes. The sequences $\mathcal{S}$ in source time window $T_1$ are highly correlated with the sequences $\mathcal{T}$ in target window $T_2$ by the effect of flow conservation.}
\label{fig_flowtensor}
\end{figure}

\textbf{Packet Flow.} An $s$-$d$ \textit{flow} is a sequence of packets encapsulating the message bits and getting forwarded from source $s$ along an assigned routing path $P\in \mathbb{Z}^L$ to destination $d$. Let $x_{f,n}^t\in \mathbb{R}$
be the value of feature $f$ measured at node $n\in P$ during the time interval $t$. For instance, $x_{f,n}^t$ can be the average packet travelling delay or transmission/receiving rate of a flow. Let us denote the time-series tensor
of a flow as $X^t=\{x_n^t\}_{n=1,\cdots,L} \in \mathbb{R}^{L\times N_f}$. With telemetry monitors installed at each node, we can collect the running traces of $X^t$ as shown in  Fig.~\ref{fig_flowtensor_a}.

\noindent\textbf{Packet Flow Prediction Problem.} Given the time-series tensor $X^t$ of a flow in the form of Fig.~\ref{fig_flowtensor_a}, predict its future service qualities $x_{f,n}^{t+1}$, such as next-time-step delay or transmission/receiving rate.

\noindent\textbf{Congestion Control.} A congestion control model can be regarded as mapping a locally-perceived history of feedback from the receiver, which reflects past traffic and network conditions, to the next choice of sending rate \cite{Nathan_Congestion_Control2019}. Such built-in controls in today's networking systems can avoid congestive collapse arising from overloading.

\noindent\textbf{Flow Conservation.}
For a packet flow with a given routing path, we have the following conservation constraint\footnote{Packets may be lost when congestion occurs. With congestion control models and buffering mechanisms, the loss probability is quite small ($<0.1\%$) in current commercial networking systems and can be tolerated by the statistical data learning process.}:
\begin{equation}\label{flowconservlaw}
\sum_{t}x^t_{r,n} = \sum_{t}x^t_{r,n+1} = b_n, \forall (n,n+1)\subseteq P
\end{equation}
where $b_n$ are the cumulative number of flow bits the source sends out.

Eq.~\ref{flowconservlaw} analytically constrains the sending and receiving bit rates at each node along the routing path. As a consequence, without congestion, the bit rates observed at each routing node will be exactly the same. Nevertheless, if routing nodes experience different congestion due to link intersection, the observed rates between two neighboring routing nodes will become larger and smaller over time during the flow lifespan. As shown in Fig.~\ref{fig_flowtensor_b}, this explicitly forms \textbf{\textit{a set of paired source-target time windows}} within the time series of two neighboring nodes. During the source window $T_1$, a node (e.g., node 1) presents larger rates than its successor node (node 2). Then, in the target window $T_2$, node 1 will present smaller rates than node 2. Finally, the cumulative amount of forwarded flow bits are conserved among the routing nodes. Such property arises from the routing structure and packet buffering mechanism in communications networks, and is independent of network and flow configurations.

Accurately modelling the flow dynamics requires attention to the above domain-specific physics and properties, which are absent in most (if not all) existing learning models.

Next, we introduce FlowNN to embed the physics as a learning bias to improve data analysis.

\noindent\textbf{Remark 1:} We are not resorting to learning a strict conservation constraint with FlowNN. Instead, we introduce the paired source-target windows as a \textit{soft constraint} or \textit{relational inductive bias} \cite{InductiveBias2018} to improve predictions by referring only to the spatio-temporal data filtered by each paired window. For example, in Fig.~\ref{Packet_flows_in_IP_networks}, $x_{r,n_0}^{t_1}=1Gbps$ is helpful for predicting $x_{r,R_1}^{t_2}=1Gbps$, while $x_{r,R_1}^{t_1}=0Gbps$ is noisy information and thus should be excluded for such prediction.

\section{Flow Neural Network}

\begin{table}[t]
\footnotesize
\centering
\caption{Notation of key symbols}
\label{symbolNotation}
\begin{tabular}{|p{1.8cm}p{5.8cm}|}
\hline
$L,N_f,d,N$ & length of a routing path $P$, number of features, number of hidden dimensions and number of recurrent iterations in FlowNN, respectively\\ \hline
$x_{f,n}^t\in \mathbb{R}$ & value of feature $f$ at node $n\in P$, and $x_{r,n}^t$ specifically denotes the transmit/receive rates  \\ \hline
$x_n^t \in \mathbb{R}^{N_f}$ & a set including all $N_f$ features at $n$ and $t$\\ \hline
$X^t \in \mathbb{R}^{L\times N_f}$ & time-series tensor of a flow at $t$\\ \hline
$\hat{h}_n^t\in \mathbb{R}^d$ & hidden vector of node $n$ at $t$\\ \hline
$f_{L_1} $ & neural net for initial embedding of $x_n^t$\\ \hline
$f_{L_2} $ & compound neural net with a multi-layer perceptron (MLP) and a recurrent net to aggregate all $h_n^t$ along dimension-$n$ and then dimension-$t$\\ \hline
$f_{L_3}$ & compound neural net with a MLP to aggregate the states of two neighboring nodes at aligned $t$, and a Seq2Seq net to predict the target window from its correlated source window\\
\hline
\end{tabular}
\end{table}

\begin{figure}[t]
\centering
\includegraphics[width=.99\columnwidth]{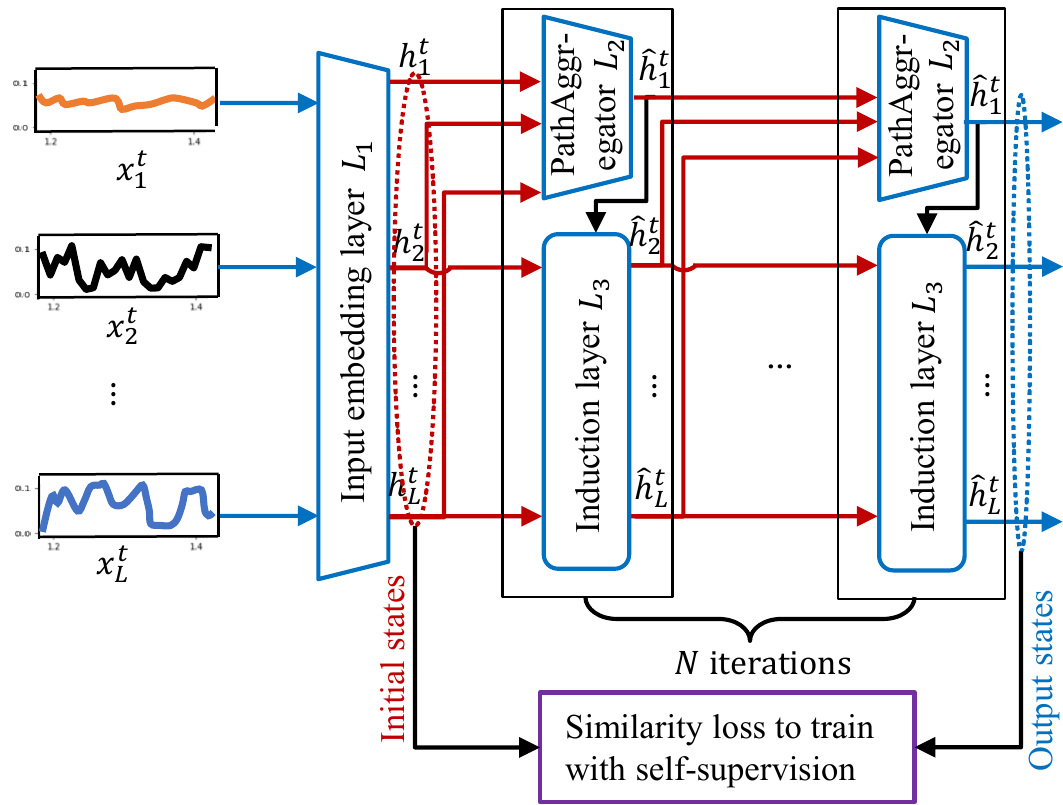} 
\caption{FlowNN architecture.}
\label{FlowNN_Framework}
\end{figure}

Our architecture (Fig.~\ref{FlowNN_Framework}) consists of three learnable neural network functions, denoted as $f_{L_1}$, $f_{L_2}$, $f_{L_3}$, respectively. FlowNN works upon the initial embedding layer $L_1$ to further impose the learning bias from the working mechanism of physical systems.

Concretely, for the time-series tensor of each flow $X^t=\{x_n^t\}_{n=1,\cdots,L} \in \mathbb{R}^{L\times N_f}$, $f_{L_1}$ use a MLP to project each $x_n^t\in \mathbb{R}^{N_f}$ to latent space $\mathbb{R}^d$:
\begin{equation}\label{pathaggregator}
h_n^\tau=f_{L_1}(x_n^\tau), \forall \tau=t_0,\cdots,t; n=1,\cdots,L
\end{equation}

Then, PathAggregator $f_{L_2}$ uses a MLP to aggregate $H^\tau=\{h_n^\tau\}_{n=1,\cdots,L}\in \mathbb{R}^{L\times d}$ from layer $L_1$ along dimension-$n$ and a recurrent net (e.g., GRU) to aggregate over time $\tau$:
\begin{equation}\label{pathaggregator}
\hat{h}_1^\tau=f_{L_2}(H^\tau)=GRU(MLP(H^\tau))\in \mathbb{R}^{d}
\end{equation}

The output of PathAggregator, $\hat{h}_1^\tau$, aggregates the locally-perceived traffic and network conditions along the routing path, which can be mapped to its next sending rate from the source node. It physically plays the role of a simulator to imitate the decision process of existing congestion control models.

With the aggregated path state $\hat{h}_1^t$ and the local hidden states of individual nodes, $\{h_n^t\}_{n=2,\cdots,L}$ from $f_{L_1}$, the induction layer $L_3$ is then applied to estimate the new local states of each node under the new sending state (i.e., $\hat{h}_1^\tau$). Specifically, for each pair\footnote{$\langle\cdot\rangle$ denotes the paired set of the two included elements.} of two neighboring nodes $\langle n,n+1\rangle$, $n=1,\cdots,L-1$, we split the associated time-series $\hat{h}_n^\tau$ into a set of correlation windows $\langle T_1^i,T_2^i\rangle_{i=1,2,\cdots}$ such that the boundaries of each time window correspond to timestamps where the rate difference between two neighboring nodes is 0, i.e., $x_{r,n}^\tau-x_{r,n+1}^\tau = 0$ (see Fig.~\ref{fig_flowtensor_b}). For each correlated window $\langle T_1^i,T_2^i\rangle$, we encode the hidden states of sequences in $T_1^i$ and $T_2^i$ as follows:
\begin{eqnarray}
\hat{h}^\tau_{n+1}=MLP\big{(}\{\hat{h}_{1}^{\tau}~||~h_{n+1}^{\tau}\}\big{)}\in\mathbb{R}^d, \forall \tau \in T_1^i \label{eqb} \\
\hat{h}_{n+1}^\tau=Seq2Seq\big{(}h_{n+1}^{\tau-1}~ | ~\{\hat{h}_1^{t}\}_{t\in T_1^i}\big{)}\in\mathbb{R}^d \label{eqc}, \forall \tau\in T_2^i
\end{eqnarray}
where $Seq2Seq$ is a sequence to sequence net \cite{S2S2014} to predict the target sequences $\{\hat{h}_{n+1}^\tau\}_{\tau\in T_2^i}$ conditioned on the states of source sequences $\{\hat{h}_1^{t}\}_{t\in T_1^i}$. In the experiments, we used both GRU-type encoders and decoders to implement the $Seq2Seq$ net. Here, $||$ and $|$ are concatenation and conditional operators, respectively.

Eq.~\ref{eqb} introduces path-level messages (i.e., $\hat{h}_1^\tau$) into the locally-perceived states of node $n+1$ for all $\tau \in T_1^i$. Eq.~\ref{eqc} predicts the states of sequences in $T_2^i$ at $n+1$ conditioned on the states of history window $T_1^i$.

The steps from Eq.~\ref{pathaggregator}--\ref{eqc} explicitly embed the natural data connections in Fig.~\ref{fig_flowtensor_a} arising from the congestion control and hop-by-hop data transmission. As shown in Fig.~\ref{FlowNN_Framework}, the predicted outputs $\{\hat{h}_n^t\}_{n=1,\cdots,L}$ can be recurrently fed into the same PathAggregator $L_2$ and Induction $L_3$ to propagate messages by the rules of Eq. \ref{pathaggregator}--\ref{eqc} to farther distances. The number $N$ of recurrent iterations is a tunable hyperparameter.

\section{Self-Supervised Training Strategy}
\begin{figure}[t]
\centering
\includegraphics[width=.999\columnwidth]{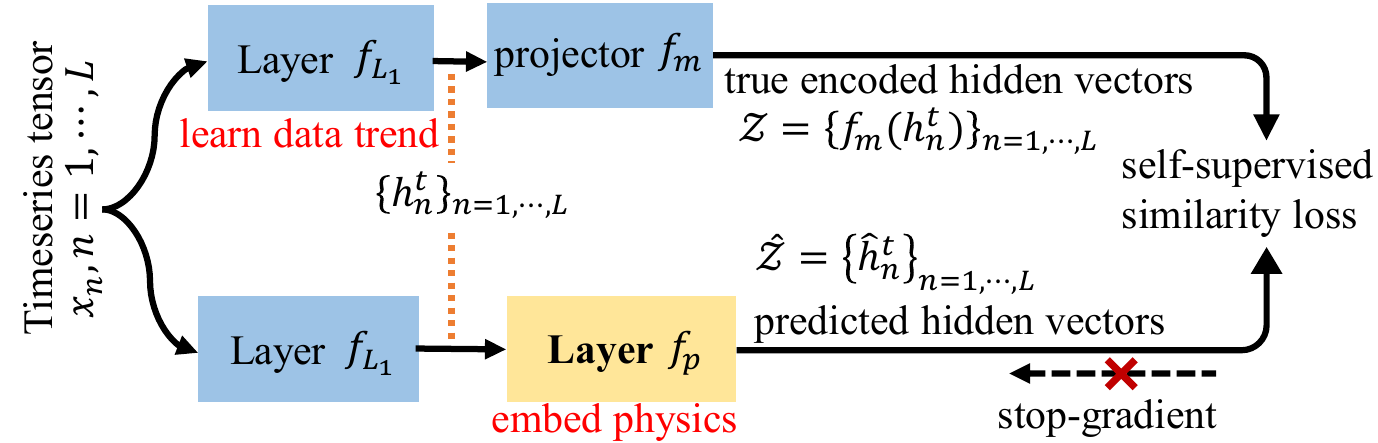} 
\caption{Self-supervised FlowNN training architecture.}
\label{SelfSuperFlowNN_Framework}
\end{figure}

The self-supervised FlowNN training architecture (Fig.~\ref{SelfSuperFlowNN_Framework}) has two branches, working on a shared initial embedding layer (encoder) $f_{L_1}$. The bottom branch predicts the states at time $t$ based on the information up to time $t-1$ by using $f_p$ (i.e., PathAggregator $L_2$ and Induction $L_3$) to impose the physical bias upon the outputs of $f_{L_1}$. The top branch embeds the ground truth features at time $t$ with a MLP projection $f_m$ that transforms the outputs from $f_{L_1}$. Denote the output vectors of the two branches as $\mathcal{Z}=(f_m(h_n^{t}))_{n=1,\cdots,L}$ and $\hat{\mathcal{Z}}=\{\hat{h}_n^{t}\}_{n=1,\cdots,L}$. Similar to \cite{BYOL2020,SimpleSiamese2021}, we define the following symmetrized loss to minimize their negative cosine similarity:
\begin{equation}\label{cosLoss}
\mathcal{L}=-{\mathrm{cos}(\mathrm{stopgrad}(\hat{\mathcal{Z}}),\mathcal{Z})\over{2}}-{\mathrm{cos}(\hat{\mathcal{Z}},\mathrm{stopgrad}(\mathcal{Z}))\over{2}}
\end{equation}
where the $\mathrm{stopgrad}(\hat{\mathcal{Z}})$ operation detaches $\hat{\mathcal{Z}}$ from the gradient computation, and thus $f_p$ receives no gradient from the first term of loss $\mathcal{L}$ (and vice versa for $\mathcal{Z}$).

The training strategy in Fig.~\ref{SelfSuperFlowNN_Framework} improves Contrastive Predictive Coding \cite{CPC2019} with a stop-gradient operation. As annotated in Fig.~\ref{SelfSuperFlowNN_Framework}, the output of the bottom pipeline is the predicted hidden vectors (the information at $t$ is NOT included when predicting for time $t$ with $f_p$). If we first remove the projector $f_m$, the upper pipeline produces the true encoded hidden vectors from $f_{L_1}$ with the information at $t$ included (thus this can be treated as the ground truth of hidden vectors in latent space). The loss here is constructed to make the predicted hidden vectors agree with the true encoded hidden vectors. The pretraining process tries to learn with FlowNN the physical bias arising from the built-in mechanisms of networks. Here, the $f_m$ is added to approximate the expectation of all possible inputs from each given input sample. This makes the learned physical bias from FlowNN stable across different inputs. This precisely explains the good transferability of the pretrained FlowNN in cross-task tests (Table 3) and Out-Of-Distribution tests (Table 4). Refer to Appendix C for more details.

\section{Experiments}
We perform experiments to predict the end-to-end packet transmission delay and sending/receiving rates $x_{r,n}^{t+1}$ along the routing path of each individual flow. The predictions of these two features can provide critical information for the transmission optimization for each service.

\noindent\textbf{Datasets.} Four publicly available datasets--\textit{WIDE}, \textit{NUMFabric} \cite{numfabric2016}, \textit{NSF} and \textit{GM50} \cite{GM50} are used.
\textit{WIDE} consists of realistic traffic traces collected from real-world network environments \cite{WIDEdataset}, and the other datasets consist of synthetic traffic widely used in recent networking research \cite{numfabric2016,GCNGAN2019}. We sampled 50 flows at the timescale of $1 ms$ for a total length of $30000 ms$. The raw features included in the time-series tensor of each flow are the sending/receiving rates, as well as the associated aggregated rates of all background flows\footnote{At each routing node of a flow, all other flows traversing the same node are in the background of the given flow.} at each routing node. Each flow has a routing path with 5 nodes in \textit{WIDE} and \textit{NUMFabric}, and variable-length routing paths ranging from 2 to 7 in \textit{NSF} and \textit{GM50}. In all experiments, we split the time series of 40 flows to train, validate and test by ratio 6:2:2, and the remaining 10 flows are retained to test the performance of the model on unseen flows after training.

\noindent\textbf{Baselines.} We have compared FlowNN with the following most related baselines: 1) Auto-Regression Integrated Moving Average (ARIMA) \cite{SARIMA2006}; 2) GRU \cite{GRU2014}; 3) multivariate GRU (m-GRU); 4) STHGCN \cite{Kalander2020}, and 5) STGCN \cite{GCNGRU2018}. Particularly, GRU encodes and predicts each time-series sequence independently without any reference to the information from other spatial nodes. In contrast, m-GRU processes joint information from all routing nodes to predict. STHGCN is the very recent work for networking traffic analysis, which uses graph convolutional networks \cite{GCN2017} and GRUs to predict the spatial states of all nodes at each time. Finally, STGCN is a widely used model for transportation traffic analysis, which is built upon the spatio-temporal graph convolution net.

The compared baselines treat the time-series tensor of a flow as multivariate input (GRU, m-GRU) or graph-type input (STHGCN, STGCN), which are the dominant approaches to model the spatio-temporal relationships in the existing studies. Note that FlowNN reduces to m-GRU model if we remove the induction layer $L_3$ and set the number of iterations to $N=1$ in Fig.~\ref{FlowNN_Framework}. Therefore, we did not perform further ablation studies in the following experiments. We conducted a grid search over the hyper-parameters for all baselines. The recurrent iteration $N=2$ was chosen. Experiments were conducted on a single Nvidia P40 24GB GPU. More details about the model implementations and hyper-parameter search procedures are given in Appendix E.

\noindent\textbf{Evaluation metrics.}
We used the following metrics to evaluate the prediction quality of $\hat{y}$ with the ground truth $y$.
\begin{equation}\label{MSELoss}
\text{Mean squared error: }\quad\quad MSE = {1\over{n}}\sum_{i=1}^n(y_i-\hat{y}_i)^2
\end{equation}
\begin{equation}
\text{Relative squared error: \ }\quad RSE = {\sum_{i=1}^n(y_i-\hat{y}_i)^2\over{\sum_{i=1}^n(y_i-\bar{y})^2}}
\end{equation}
\text{~~Correlation coefficient:}
\begin{equation}
Corr = {\sum_{i=1}^n(y_i-\bar{y})(\hat{y}_i-\hat{\bar{y}})\over{\sqrt{\sum_{i=1}^n{(y_i-\bar{y}_i)^2}}\sqrt{\sum_{i=1}^n{(\hat{y}_i-\hat{\bar{y}})^2}}}}
\end{equation}
where $\bar{y}$ and $\hat{\bar{y}}$ are the mean values of $y$ and $\hat{y}$.

\begin{table*}[t]
\footnotesize
\centering
\caption{Test performance comparisons}
\label{table_step1Predict}
\begin{tabular}{|p{.1cm}|p{1.3cm}|p{2.15cm}p{2.15cm}p{2.15cm}|p{2.15cm}p{2.15cm}p{2.15cm}|}
\hline
                 \multicolumn{2}{|c|}{}                         &                               \multicolumn{3}{c|}{ Test on seen flows}      &       \multicolumn{3}{c|}{ Test on unseen flows}           \\ \cline{3-8}
\multicolumn{2}{|c|}{Dataset$|$Model}                          &MSE                            & RSE                    & Corr               &MSE                            & RSE                    & Corr\\  \hline\hline
\multirow{4}{*}{\rotatebox[origin=c]{90}{\textit{NUMFabric\quad\ \ }}}   &    \textit{Naive}   &  0.3567   & 0.6175    & 0.8091         &  \quad-- & \quad--  & \quad-- \\
                 &    \textit{ARIMA}        &  0.2724 &  0.5199 & 0.8543               &  \quad-- & \quad--  & \quad--  \\
                 &    \textit{GRU}         &  0.0872$\pm$0.0013 & 0.3021$\pm$0.0018 & 0.9533$\pm$0.0006 & 0.0895$\pm$0.0033 & 0.3092$\pm$0.0038 & 0.9510$\pm$0.0013 \\
                 &    \textit{m-GRU}        &  \underline{0.0306$\pm$0.0004} & \underline{0.1789$\pm$0.0013} & \underline{0.9840$\pm$0.0002} &  \underline{0.0311$\pm$0.0010} & \underline{0.1823$\pm$0.0025} & \underline{0.9833$\pm$0.0005} \\
                 &    \textit{STHGCN}        &  0.0405$\pm$0.0006 & 0.2060$\pm$0.0013 & 0.9785$\pm$0.0003 & 0.0417$\pm$0.0014 & 0.2109$\pm$0.0026 & 0.9775$\pm$0.0006 \\
                 &    \textit{STGCN}        &  0.0457$\pm$0.0007 & 0.2186$\pm$0.0018 & 0.9758$\pm$0.0006 & 0.0453$\pm$0.0012 & 0.2200$\pm$0.0039 & 0.9755$\pm$0.0018 \\
                 &    \textbf{\textit{FlowNN}}        &  \textbf{0.0254$\pm$0.0003} & \textbf{0.1632$\pm$0.0010} & \textbf{0.9867$\pm$0.0002} & \textbf{0.0261$\pm$0.0009} & \textbf{0.1672$\pm$0.0021} & \textbf{0.9860$\pm$0.0004} \\
                  \hline
\multirow{4}{*}{\rotatebox[origin=c]{90}{\textit{WIDE\quad\quad\quad}}}   &    \textit{Naive}   &  0.8384   & 0.9420     & 0.5533         &  \quad-- & \quad--  & \quad-- \\
                 &    \textit{ARIMA}        &  0.4879 & 0.6422 & 0.7668               &  \quad-- & \quad--  & \quad--  \\
                 &    \textit{GRU}         &  0.5857$\pm$0.0074 & 0.7764$\pm$0.0057 & 0.6325$\pm$0.0071 & 0.5508$\pm$0.0113 & 0.7539$\pm$0.0088 & 0.6585$\pm$0.0103 \\
                 &    \textit{m-GRU}        &  \underline{0.4609$\pm$0.0061} & \underline{0.6887$\pm$0.0059} & \underline{0.7271$\pm$0.0055} & \underline{0.4490$\pm$0.0087} & \underline{0.6807$\pm$0.0086} & \underline{0.7345$\pm$0.0080} \\
                 &    \textit{STHGCN}        &  0.4878$\pm$0.0063 & 0.7085$\pm$0.0056 & 0.7079$\pm$0.0055 & 0.4710$\pm$0.0096 & 0.6971$\pm$0.0084 & 0.7191$\pm$0.0084 \\
                 &    \textit{STGCN}        &  0.4967$\pm$0.0021 & 0.7151$\pm$0.0013 & 0.7073$\pm$0.0014 & 0.4695$\pm$0.0048 & 0.6961$\pm$0.0032 & 0.7258$\pm$0.0034 \\
                 &    \textbf{\textit{FlowNN}}        &  \textbf{0.4123$\pm$0.0054} & \textbf{0.6514$\pm$0.0058} & \textbf{0.7599$\pm$0.0050} & \textbf{0.4007$\pm$0.0078} & \textbf{0.6430$\pm$0.0085} & \textbf{0.7674$\pm$0.0073} \\
                  \hline
\multirow{4}{*}{\rotatebox[origin=c]{90}{\textit{NSF\quad\quad\quad}}}   &    \textit{Naive}   &  0.2244   & 0.5006      & 0.8742        &  \quad-- & \quad--  & \quad-- \\
                 &    \textit{ARIMA}        &  0.2240 & 0.4852 & 0.8745               &  \quad-- & \quad--  & \quad--  \\
                 &    \textit{GRU}         &  0.1168$\pm$0.0030 & 0.3576$\pm$0.0046 & 0.9341$\pm$0.0018 & 0.1140$\pm$0.0047 & 0.3814$\pm$0.0052 & 0.9253$\pm$0.0021 \\
                 &    \textit{m-GRU}        &  \underline{0.0794$\pm$0.0022} & \underline{0.2948$\pm$0.0041} & \underline{0.9556$\pm$0.0013} & \underline{0.0888$\pm$0.0040} & \underline{0.3366$\pm$0.0050} & \underline{0.9423$\pm$0.0018} \\
                 &    \textit{STHGCN}        & 0.0938$\pm$0.0024 & 0.3204$\pm$0.0042 & 0.9474$\pm$0.0014 & 0.1302$\pm$0.0065 & 0.4075$\pm$0.0071 & 0.9157$\pm$0.0028 \\
                 &    \textit{STGCN}        &  0.1214$\pm$0.0016 & 0.3643$\pm$0.0027 & 0.9333$\pm$0.0013 & 0.1222$\pm$0.0034 & 0.3944$\pm$0.0055 & 0.9213$\pm$0.0035 \\
                 &    \textbf{\textit{FlowNN}}        & \textbf{0.0791$\pm$0.0021} & \textbf{0.2942$\pm$0.0039} & \textbf{0.9558$\pm$0.0012} & \textbf{0.0851$\pm$0.0035} & \textbf{0.3296$\pm$0.0043} & \textbf{0.9446$\pm$0.0015} \\
                  \hline
\multirow{4}{*}{\rotatebox[origin=c]{90}{\textit{GM50\quad\quad\quad}}}   &    \textit{Naive}   &  1.3387   & 1.1929     & 0.2806         &  \quad-- & \quad--  & \quad-- \\
                 &    \textit{ARIMA}        &  \textbf{0.5383} & \underline{0.7687} & \underline{0.6410}               &  \quad-- & \quad--  & \quad--  \\
                 &    \textit{GRU}         &  0.6381$\pm$0.0151 & 0.8179$\pm$0.0113 & 0.5756$\pm$0.0164 & 0.8921$\pm$0.0116 & 0.9494$\pm$0.0064 & 0.3148$\pm$0.0274 \\
                 &    \textit{m-GRU}        & 0.5628$\pm$0.0143 & 0.7682$\pm$0.0116 & 0.6405$\pm$0.0134 & \underline{0.8168$\pm$0.0110} & \underline{0.9084$\pm$0.0064} & \underline{0.4184$\pm$0.0147} \\
                 &    \textit{STHGCN}       & 0.5853$\pm$0.0149 & 0.7834$\pm$0.0117 & 0.6218$\pm$0.0143 & 0.8427$\pm$0.0123 & 0.9227$\pm$0.0076 & 0.3866$\pm$0.0175 \\
                 &    \textit{STGCN}        & 0.5877$\pm$0.0021 & 0.7850$\pm$0.0017 & 0.6206$\pm$0.0023 & 0.8329$\pm$0.0051 & 0.9175$\pm$0.0020 & 0.3997$\pm$0.0040 \\
                 &    \textbf{\textit{FlowNN}}       & \underline{0.5429$\pm$0.0138} & \textbf{0.7544$\pm$0.0114} & \textbf{0.6567$\pm$0.0125} & \textbf{0.7958$\pm$0.0106} & \textbf{0.8967$\pm$0.0061} & \textbf{0.4431$\pm$0.0131} \\
                  \hline
\end{tabular}
\end{table*}

\subsubsection{Linear evaluation on rate prediction task.}
We first evaluate the learned latent representation $\{\hat{h}_n^t\}_{n=1,\cdots,L}$ for the rate prediction task by finetuning a MLP readout layer on top of the pretrained FlowNN model. During the finetuning process, the
pretrained FlowNN model is further optimized according to ground-truth labels. By contrast, the compared baselines are directly trained together with their MLP readout layers.

Table \ref{table_step1Predict}
reports the test performance of next-step ($1ms$) predictions. Here, the \textit{Naive} approach simply predicts the next-step value to be the same as the current value. The results of \textit{Naive} and ARIMA provide evidence that traditional
statistical time-series prediction models fail to provide reasonable performance on short-timescale prediction tasks. We can observe that FlowNN outperforms all baselines, achieving a MSE decrease of up to 71\% (GRU), 17\% (m-GRU), 36.8\% (STHGCN), and 44.4\% (STGCN).

m-GRU is similar to the function of PathAggregator in FlowNN, which also integrates the effect of congestion control in communication networks. This explains the reason behind its superiority when compared with other baselines in this task (but it does not perform well in the following cross-task learning test (Table \ref{tableDelay}) and model generality experiments (Table \ref{OODtest}).). More details and intuition are provided in
Appendix B to interpret why the involved recurrent and graph convolutional operators in these baselines will fail to capture the data correlations manifested in the time-series tensors.

\subsubsection{Transfer to end-to-end packet delay prediction task.}
\begin{table}[t]
\footnotesize
\centering
\caption{Test performances of end-to-end packet transmission delay on unseen flows}
\label{tableDelay}
\begin{tabular}{|p{.1cm}|p{.1cm}|p{2cm}p{2cm}p{1cm}|}
\hline
&       *                &MSE                            & RSE                    & Corr    \\  \hline\hline
\multirow{5}{*}{\rotatebox{90}{\textit{NUMFabric}}}& 1             & 0.1051$\pm$0.0046 & 0.3204$\pm$0.0046 & 0.9473$\pm$0.0012  \\
  &2                    & 0.0386$\pm$0.0037 & 0.1942$\pm$0.0057 & 0.9811$\pm$0.0012 \\
  &3                    & 0.0512$\pm$0.0036  &  0.2236$\pm$0.0053  &  0.9747$\pm$0.0013 \\
  &4                    & \underline{0.0346$\pm$0.0022} & \underline{0.1841$\pm$0.0107} & \underline{0.9831$\pm$0.0079} \\
  &\textbf{5}                    & \textbf{0.0336$\pm$0.0032} & \textbf{0.1811$\pm$0.0057} & \textbf{0.9836$\pm$0.0011} \\  \hline
                          \hline
\multirow{5}{*}{\rotatebox{90}{\textit{WIDE}}}& 1             & 0.4002$\pm$0.0594 & 1.0371$\pm$0.0147 & 0.0355$\pm$0.0086  \\
  &2                    & 0.3977$\pm$0.0768 & 1.0340$\pm$0.0675 & \underline{0.3968$\pm$0.0292} \\
    &3                    & \underline{0.3877$\pm$0.0558}  & \underline{1.0209$\pm$0.0270}  & 0.2013$\pm$0.0230 \\
      &4                    & 0.4335$\pm$0.0609 & 1.0817$\pm$0.0406 & 0.2148$\pm$0.0058 \\
        &\textbf{5}                    & \textbf{0.3168$\pm$0.0405}  & \textbf{0.9228$\pm$0.0735}  & \textbf{0.4571$\pm$0.0369} \\  \hline
  \multicolumn{5}{l}{* :\ \ \ 1: GRU, 2: m-GRU, 3: STHGCN, 4: STGCN, 5: FlowNN}
\end{tabular}
\end{table}

It is important to guarantee that the provided end-to-end transmission delays meet users' agreements, such as a delay of less than 50\textit{ms} for online gaming. In real-world networks, many factors may influence the delay, including dynamic traffic loads, packet congestion and queueing, etc. It is difficult to build an accurate delay model even for human experts \cite{Deepq2018,routenet2019}. As explained in \textbf{Congestion Control}, the dynamics of the past traffic and networking conditions will influence the next-time sending actions and the overall transmission delay. This makes it possible to predict the delay from the generating behaviors of flow rates.

In this task, we work on the same FlowNN model pretrained in next-step prediction tasks, and finetune a new MLP readout to predict the next-time-step transmission delay. Table \ref{tableDelay} shows the test performances of different models. Although pretrained on the dataset of packet transmission rates, FlowNN still achieves the best results on the task of predicting a physical feature that is never included in the pretraining process. This shows the transferability of the self-supervised pretraining model of FlowNN across different learning tasks.

\subsubsection{Generality test on Out-Of-Distribution dataset.}
\begin{table}[t]
\footnotesize
\centering
\caption{Out-of-distribution tests}
\label{OODtest}
\begin{tabular}{|p{.1cm}|p{2.2cm}p{2.2cm}p{2.2cm}|}
\hline
          *            &MSE                            & RSE                    & Corr    \\  \hline
1             &   0.5864$\pm$0.0123 & 0.7779$\pm$0.0085 & 0.6284$\pm$0.0108  \\
2                &   0.5232$\pm$0.0111 & 0.7347$\pm$0.0076 & 0.6785$\pm$0.0085   \\
3           &    0.5003$\pm$0.0107 & 0.7185$\pm$0.0083 & 0.6955$\pm$0.0088    \\
4                &  \underline{0.4858$\pm$0.0050} & \underline{0.7081$\pm$0.0033} & \underline{0.7100$\pm$0.0035}   \\
\textbf{5}             &    \textbf{0.3959$\pm$0.0078} & \textbf{0.6391$\pm$0.0084} & \textbf{0.7692$\pm$0.0072}     \\  \hline
  \multicolumn{4}{l}{*\quad 1: GRU, 2: m-GRU, 3: STHGCN, 4: STGCN, 5: FlowNN}
\end{tabular}
\end{table}
We further test the model generality when a FlowNN model, trained on one dataset, say \textit{NUMFabric}, is used to work in an environment that is different from the trained one. Specifically, we \textit{froze} the FlowNN model trained for the next-step prediction task on the dataset \textit{NUMFabric}, and finetuned a new MLP readout to test its prediction performance on the dataset \textit{WIDE}. For a fair comparison, we also finetuned the readout layers of the frozen baselines trained on \textit{NUMFabric} to test their performances on \textit{WIDE}. From the results in Table \ref{OODtest}, we can observe that FlowNN remains the best performer. We conjecture that the source of improvement is the physical logic properly encoded in our FlowNN design, which is possible for all the flows or environments that work according to the same principles. By contrast, m-GRU degrades remarkably and gets worse than the two graph models. This shows the good generality and robustness of FlowNN, as well as the advantage of physics modeling in our approach.

\section{Related Work}
\textbf{Traditional analytical models.} The past decades have seen numerous packet analysis models proposed to mathematically model the network dynamics \cite{Gebali2015}. For example, extensive studies use the Poisson
model to characterize the traffic by assuming that the arrival pattern between two successive packets follows a Poisson process. Considering the heavy tailed distribution and burstiness of the data-center traffic, recent work in \cite{Benson2010} models the packet arrival pattern as a log-normal process. To capture the temporal patterns and make predictions accordingly, ARIMA is exploited in \cite{Autoregress2019} to model the time-series packet traffic. These analytical models often make important assumptions. Moreover, these statistical models work at coarse timescales (e.g., hours) and assume relatively smoother traffic patterns. However, many tasks require the flow statistics at a subsecond timescale, e.g., packet forwarding/queueing configurations and real-time networking control. This implies that tasks requiring analysis models at finer-grained time scales are beyond the capability of these traditional models.

\vspace{0.04cm}
\noindent\textbf{Neural network based learning models.}
With data-driven learning, packet analysis can be done by extracting the spatio-temporal patterns within the time-series data. In this field, a wide range of existing neural networks fit this task, including Convolutional Neural Nets (CNNs, \cite{CNNdatacenter2018}), Graph Neural Nets (GNNs, \cite{routenet2019}), Recurrent Neural Nets (RNNs) as well as their variants and combinations (e.g., STHGCN \cite{Kalander2020}, STGCN \cite{GCNGRU2018}, and \cite{Deepq2018,MARLGNN2021}). The designs tailored to data-specific properties enable the success of these existing models in their dedicated domains, such as the convolutional operation to capture the spatially local correlations in CNNs, and the aggregation operation to capture correlations between adjacent links in GNNs, and so on. As systems with clear structure and working mechanisms, communication networks exhibit their own specific data properties, which are difficult to capture for the incumbent models without any modification. Moreover, existing work only targets coarse-grained timescales above minutes or even hours. Models at a sub-second granularity, such as FlowNN, require a non-trivial combination of spatio-temporal data trends, system structure and working mechanisms.

Similar to existing learning models, FlowNN uses a finetuning stage to adapt a pretrained model to various downstream tasks or environments. Nevertheless, the self-supervised learning strategy used by FlowNN is able to embed the physical bias that is universally transferable, provided that the new environment follows the same working mechanisms. The patterns manifested by the data itself may vary in different environments, but the underlying physical working mechanisms typically remain invariant as long as the new environment follows the same principles, e.g., TCP/IP.

\section{Conclusion}
In this paper, we developed a customized neural network--FlowNN with the associated self-supervised training strategy to improve the short-timescale prediction problem of information flows in communication networks. This study pioneers the design of a customized neural network and learning strategy by combining network structure and working
mechanisms. We reported state-of-the-art performances for multiple practical networking applications, which demonstrates the strength of our approach. As a new `playground' for both networking and deep learning communities, the research of network intelligence is still in its infancy. We hope this work will inspire more innovation in this field in the future.

\appendix
\section{Sending Packets of Data with Congestion Control}
\begin{figure*}[t]
\centering
\begin{minipage}[c]{.745\textwidth}
\centering
\includegraphics[width=1\columnwidth]{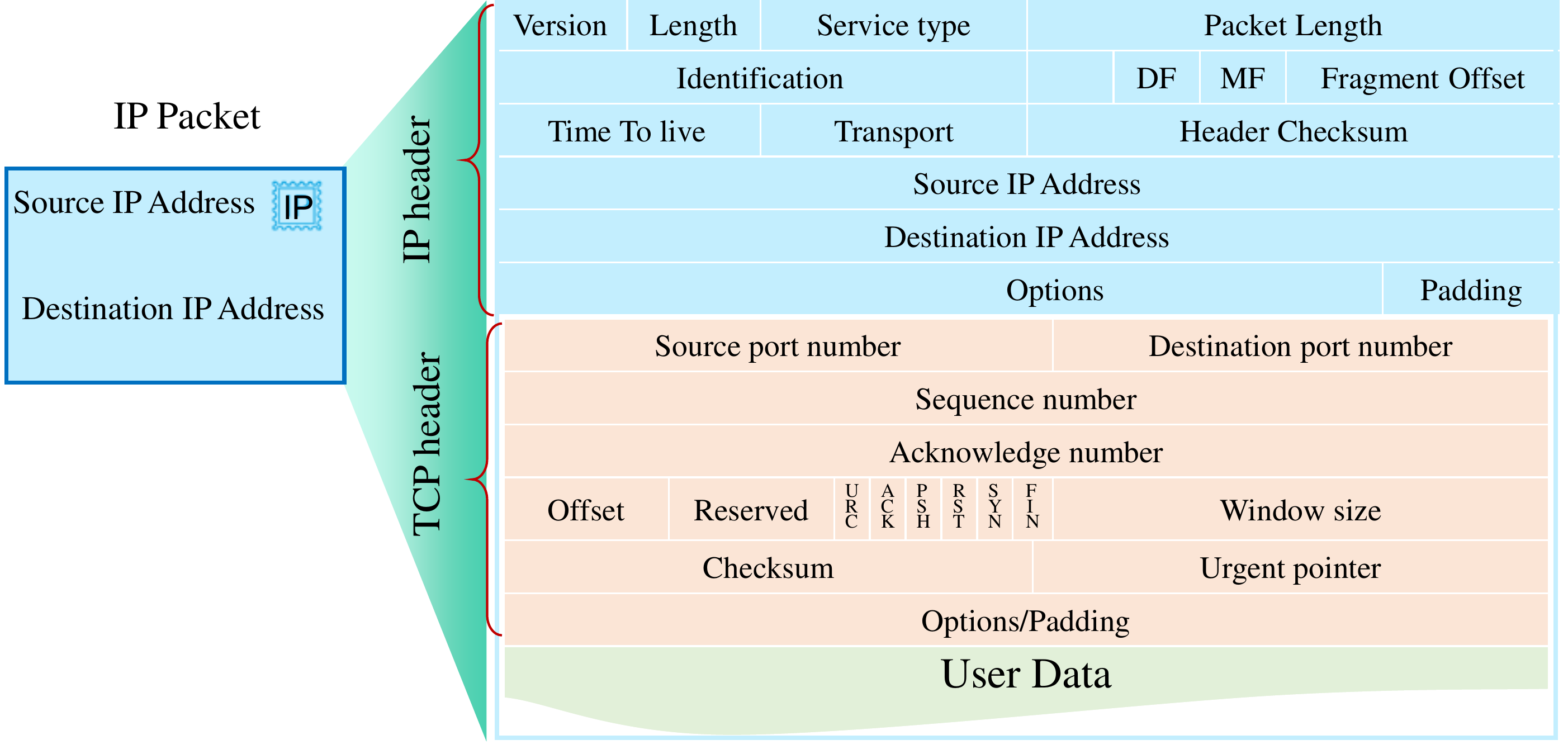} 
\caption{Encapsulation format of TCP/IP packet.}
\label{TCPIPformat}
\end{minipage}
\begin{minipage}[c]{.25\textwidth}
\centering
\includegraphics[width=.95\columnwidth]{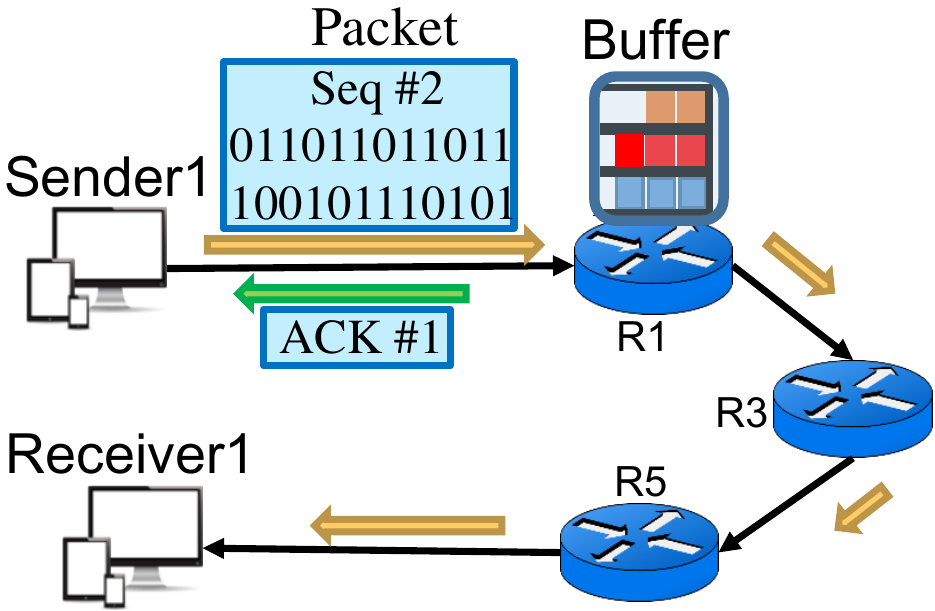} 
\caption{Data transmissions between two end hosts.}
\label{packetsending}
\end{minipage}
\end{figure*}

In a communication network with Transmission Control Protocol/Internet Protocol (TCP/IP) suits, users' messages are encapsulated into a set of small IP packets. Fig.~\ref{TCPIPformat}--\ref{packetsending} illustrate
the IP packet format and the sending process of IP packets.

\subsubsection{Congestion control} When a TCP connection (or flow) is established between two end hosts, the sender will stream data packets to its receiver, and constantly receive the acknowledgement (ACK) as feedback about the packet transmission qualities from the receiver. When the data sending rate overwhelms the link bandwidth, congestion occurs and packets will be queued in buffers or dropped at congested nodes. Congestion increases the delay to receive the packets and thus degrades the overall transmission rates.

With a congestion control model, the sender maintains a dynamic congestion window to limit the maximum number of bytes that can be sent out in response to every ACK feedback. For example, the CUBIC model \cite{Cubic2008} uses
the following growth function to determine the real-time congestion window $W(t)$:
\begin{equation}\label{Cubeq_ic}
W(t)=C(t-K)^3+W_{max}, K=\sqrt[3]{{W_{max}\beta\over{C}}}
\end{equation}
where $W_{max}$ is the window size after the last congestion event (i.e., packet loss/dropping is detected from ACK feedback); $C$ and $\beta$ are fixed constants; $t$ is the elapsed time since the last window reduction.

Eq.~\ref{Cubeq_ic} is adjusted every Round Trip Time (RTT) between the sender and receiver. Each RTT is normally a subsecond value. The quick adjustment avoids consistent congestion, but also updates the sending rate of a flow
vary frequently. The cumulative loading volume (or sending rates) of packets in the history determines the probability of packet loss. Such connection makes it possible to approximate the sending rules in Eq.~\ref{Cubeq_ic} based on
historical traces of sending rates of all flows. This explains the necessity of including the PathAggregator in FlowNN to approximate the decision process of the congestion control model.

Once the allowed sending bytes/rates are determined by the congestion control model, the sender will encapsulate the allowed number of data bytes into packets and send them out sequentially. As shown in Fig.~\ref{packetsending}, all packets will be forwarded hop-by-hop along a routing path until reaching the final destination.

\section{Relational Inductive Bias of Different Models}
Many approaches in machine learning use a \textit{relational inductive bias} \cite{InductiveBias2018} to impose constraints on relationships and interactions among entities in a learning process. Relational bias can be understood in terms of the bias-variance tradeoff \cite{biasvariance1992}, which helps improve solutions to generalize in a desirable way. However, an improper relational bias will lead to suboptimal performance. The elementary building blocks and their compositions within FlowNN and the compared baselines induce various relational inductive biases.

\subsubsection{Recurrent and convolutional layers in baselines} The popular recurrent and/or convolutional layers are used in the compared baselines. For example, with a convolutional layer, STHGCN and STGCN are actually imposing spatial locality as the inductive bias. Locality indicates correlations between entities in close proximity with one another in the input signal's coordinate space, \textit{isolated} from distant entities \cite{InductiveBias2018}. Similarly, the recurrent units in GRUs also incorporate a bias for temporal locality in the sequence. Therefore, with these building blocks, all these baselines make predictions with \textit{all entities} in both spatially and temporally close proximity in the time-series tensor. This actually forms a message propagation routine that is depicted in Fig.~\ref{relationalbiascompare}(a). That is, the messages are first aggregated along the spatial direction (i.e., vertical) and then further aggregated along the time domain (i.e., horizontal).

\subsubsection{PathAggregator and Induction layers in FlowNN} 
\begin{figure}[t]
\centering
\includegraphics[width=1\columnwidth]{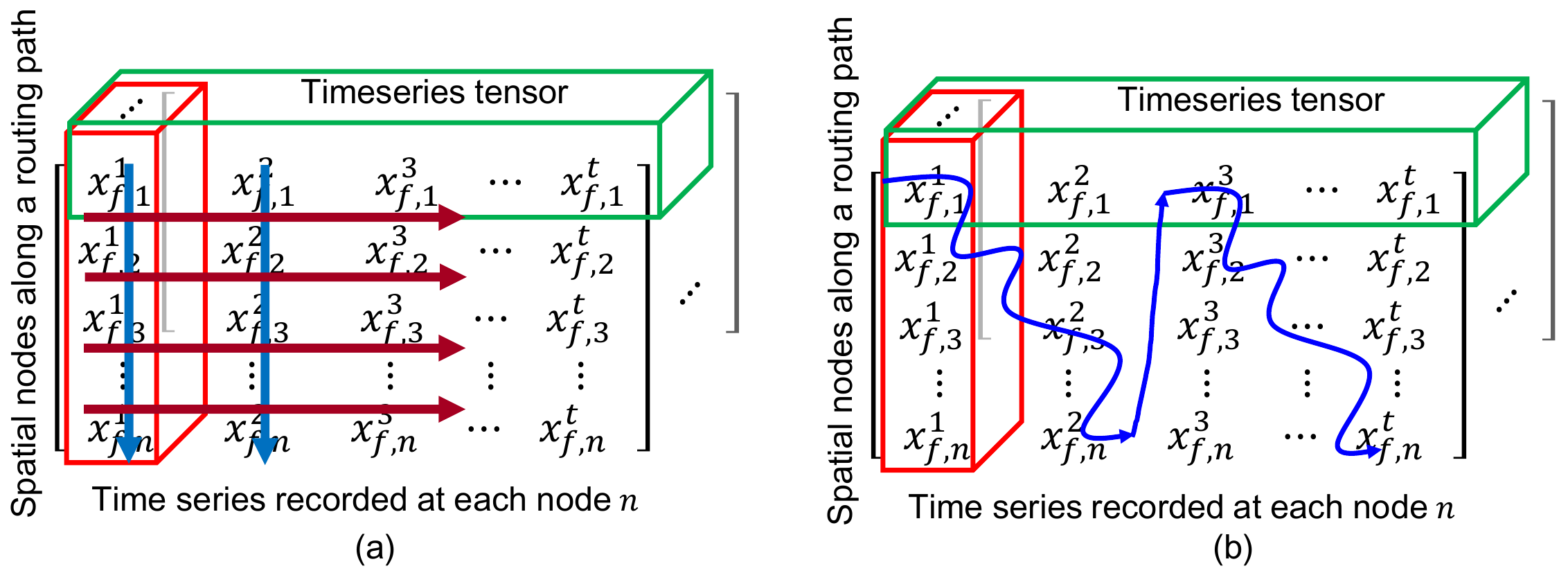} 
\caption{(a) Relational bias in baselines; (b) relational bias in FlowNN}
\label{relationalbiascompare}
\end{figure}

With PathAggregator, FlowNN only relates (predicts) the states of source node $\hat{h}_{1}^t$ with all historical information of $\{\hat{h}_{n}^{\tau}\}_{n=1,\cdots,L}$, $\tau<t$.
However, to predict states of spatial nodes $n=2,\cdots,L$, the Induction layer only relates the spatial information of $\hat{h}_{n-1}^\tau$ at its correlated time window. This filters out the irrelevant entities at the space-time coordinates that otherwise will be included as undesired noise. The blue curve in Fig. \ref{relationalbiascompare}(b) shows the routine to propagate the transmission information along the space-time coordinates.

Fig. \ref{relationalbiascompare} compares the relational biases represented by different models. FlowNN attends to the physical inductive bias that matches the natural generating behaviors of the time-series tensor.
In contrast, none of the baselines incorporate this inductive bias.

\section{Stop-Gradient for FlowNN}
\begin{figure}[t]
\centering
\includegraphics[width=.999\columnwidth]{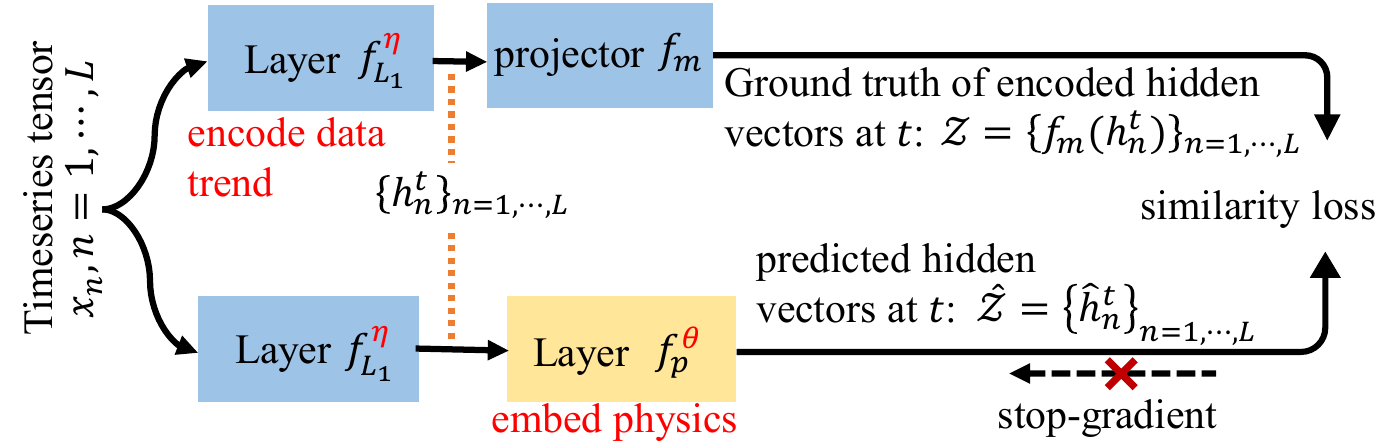} 
\caption{Self-supervised FlowNN training architecture.}
\label{SelfSuperFlowNN_Framework}
\end{figure}
Let's consider a loss function of the following form:
\begin{equation}\label{AppendCLoss}
\mathcal{L}(\eta,\theta)=\mathbb{E}_X\Big{[}\big{\|}\mathcal{F}^{(\eta,\theta)}(X)-\mathbb{E}_{X}\big{[}f_{L_1}^\eta(X)\big{]}\big{\|}_2^2\Big{]}
\end{equation}
where $\mathcal{F}^{(\eta,\theta)}$ is a FlowNN with parameters $(\eta,\theta)$ for layer $f_{L_1}^\eta$ and layer $f_p^\theta$ (i.e., $L_2$ and $L_3$) in Fig. \ref{SelfSuperFlowNN_Framework}; $X$ is a sample of the input time-series tensor.

Eq. \ref{AppendCLoss} is equivalent to the negative cosine similarity loss in the main text if all vectors are $l_2$-normalized. With this formulation, the training process updates FlowNN at each step so that the embedded physical logics by FlowNN remains stable for all possible inputs $X$. This is optimized by solving the following problem w.r.t $\eta$ and $\theta$:
\begin{equation}\label{twostepproblem}
\min_{\eta, \theta} \mathcal{L}(\eta,\theta)
\end{equation}

The variable $\theta$ is learnt in FlowNN to make predictions based on physical logics. The variable $\eta$ is optimized to represent the data trending information manifested within the input $X$ by $f_{L_1}$. Formally, the problem in Eq. \ref{twostepproblem} can be
solved by alternatively solving the following two subproblems:
\begin{eqnarray}
\eta^t\leftarrow \arg\min_{\eta} \mathcal{L}(\eta, \theta^{t-1})\label{eqsubproblem1}\\
\theta^t\leftarrow \arg\min_{\theta} \mathcal{L}(\eta^t,\theta) \label{eqsubproblem2}
\end{eqnarray}
where $t$ is the index of alternation and $\leftarrow$ means assigning.

Subproblem (\ref{eqsubproblem1}) attempts to optimize the representation towards the input under a constant $f_p^{\theta^{t-1}}$. The \textit{stop-gradient operation} in Fig. \ref{SelfSuperFlowNN_Framework} is a natural implementation, since the gradient does not back-propagate to $\theta^{t-1}$ (a constant) in this subproblem. By definition of Eq. \ref{AppendCLoss}, $\eta^t$ is expected to minimize the average representations over any possible input $X$. In practice, it is unrealistic to actually compute the expectation $\mathbb{E}_{X}$. Similar to \cite{SimpleSiamese2021}, we alternatively apply a neural network $f_m$ (a MLP projection in the experiments) in Fig. \ref{SelfSuperFlowNN_Framework} to predict the expectation. With the second term of the defined symmetrized loss in the main text, subproblem (\ref{eqsubproblem2}) will be solved to optimize the learned physical bias from FlowNN. Therefore, the two subproblems can be solved by one-step gradient update through such training structure.

Although the above analysis is a similar application of stop-gradient operation as \cite{SimpleSiamese2021,BYOL2020}, we want to highlight that the self-supervised training strategy for FlowNN is not a standard application of \cite{SimpleSiamese2021}. Instead, we combined the advances of Contrastive Predictive Code (CPC) \cite{CPC2019} with a stop-gradient operation. More importantly, the goal of the designed loss is to improve the prediction quality by comparing the predicted hidden vectors with the ground-truth labels in latent space. The projector $f_m$ receives as input the encoded information at $t$, which however is not seen by Layer $f_p^\theta$ when generating the predicted hidden vectors at $t$. Moreover, the projector $f_m$ is a MLP, while $f_p^\theta$ contains GRU-type calculations. Consequently, $f_m$ and $f_p^\theta$ will produce different outputs even when $f_{L_1}^\eta$ becomes a constant mapping. This implies that a constant mapping of $f_{L_1^\eta}$ will not lead to the minimization of the similarity loss. Therefore, there is no learning collapsing problem in the self-supervised learning architecture of Fig. \ref{SelfSuperFlowNN_Framework}. This work is significantly different from the solution to the collapsing problem and the goal of the Siamese design in \cite{SimpleSiamese2021}, which aims at extracting similar features from two augmented views.
 \section{Extra Experimental Results}
 \begin{figure}[t]
\centering
\begin{subfigure}{.5\columnwidth}
\centering
\includegraphics[width=.98\columnwidth]{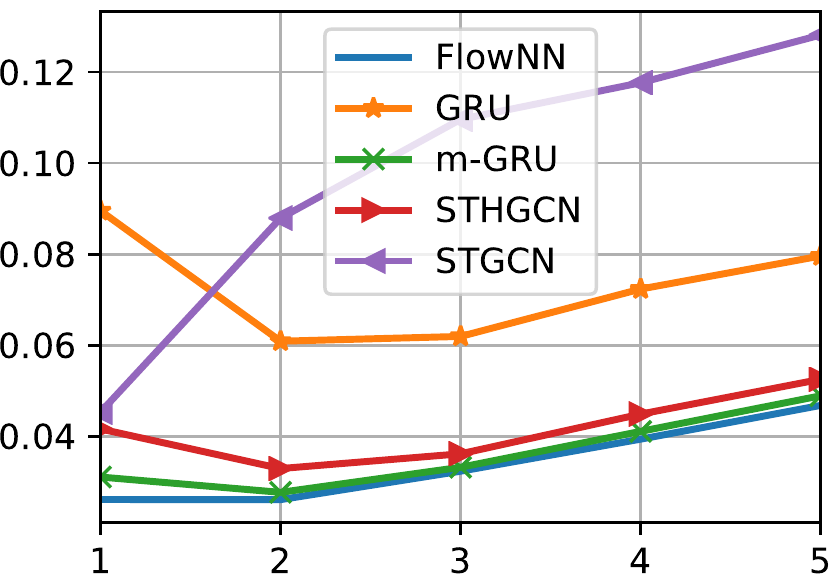}
\caption{Test MSE on \textit{NUMFabric}}
\label{fig77}
\end{subfigure}%
\begin{subfigure}{.5\columnwidth}
\centering
\includegraphics[width=.98\columnwidth]{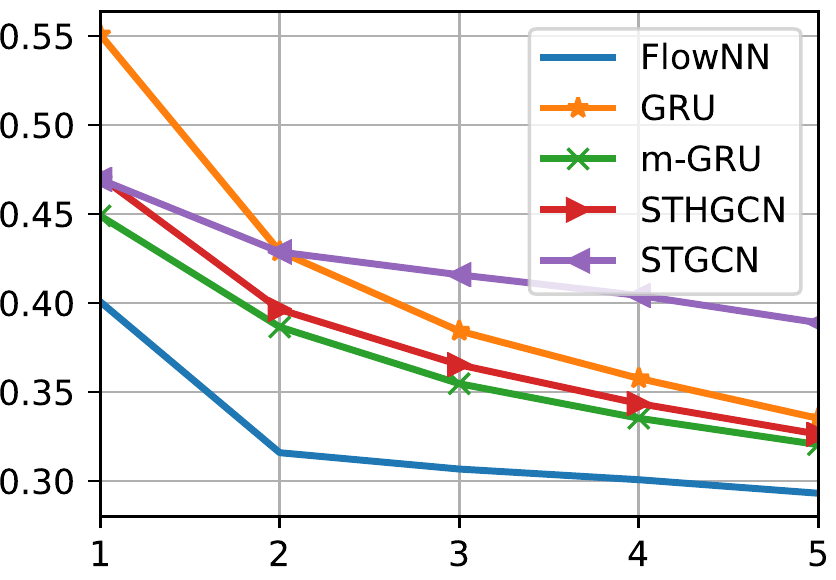}
\caption{Test MSE on \textit{WIDE}.}
\label{fig79}
\end{subfigure}%
\newline
\begin{subfigure}{.5\columnwidth}
\centering
\includegraphics[width=.98\columnwidth]{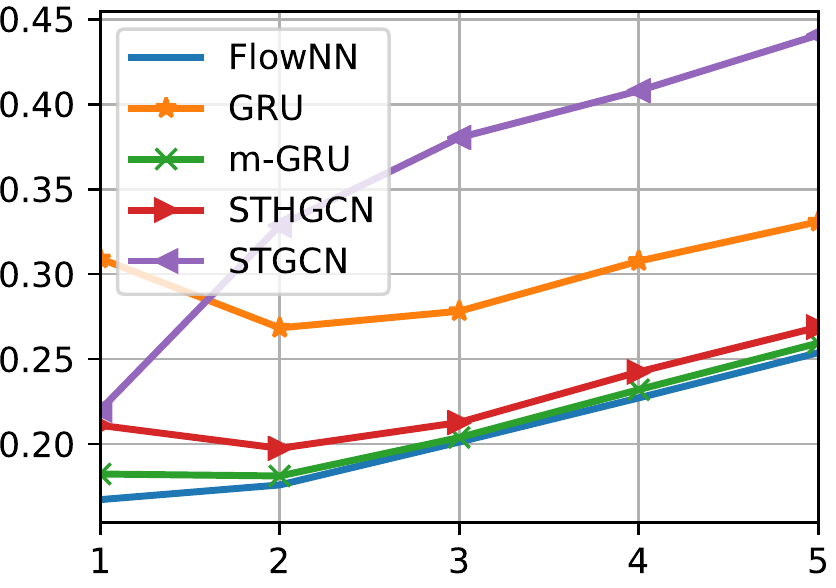}
\caption{Test RSE on \textit{NUMFabric}.}
\label{fig90}
\end{subfigure}%
\begin{subfigure}{.5\columnwidth}
\centering
\includegraphics[width=.98\columnwidth]{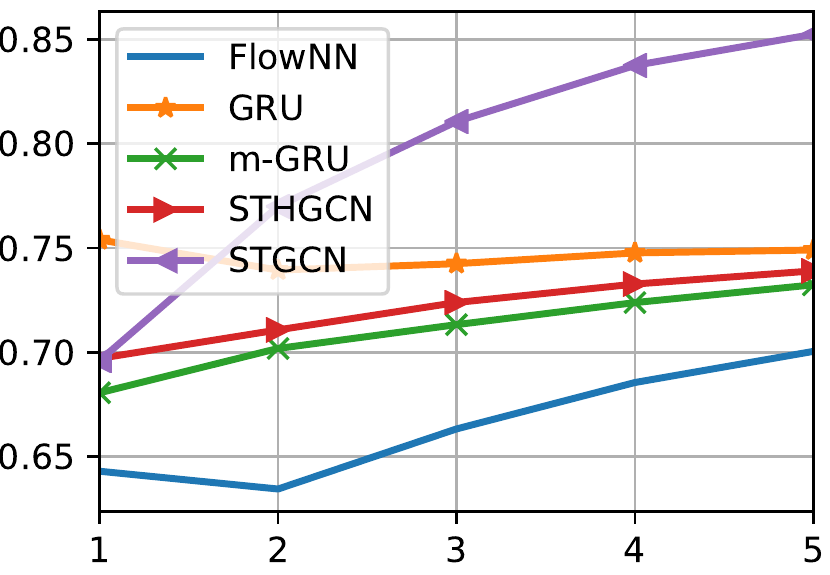}
\caption{Test RSE on \textit{WIDE}.}
\label{fig77}
\end{subfigure}%
\newline
\begin{subfigure}{.5\columnwidth}
\centering
\includegraphics[width=.98\columnwidth]{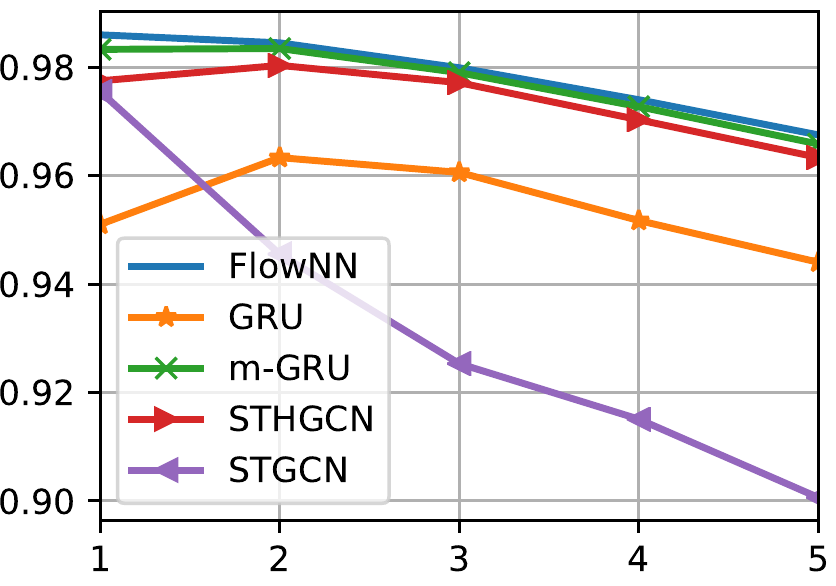}
\caption{Test Corr on \textit{NUMFabric}.}
\label{fig79}
\end{subfigure}%
\begin{subfigure}{.5\columnwidth}
\centering
\includegraphics[width=.98\columnwidth]{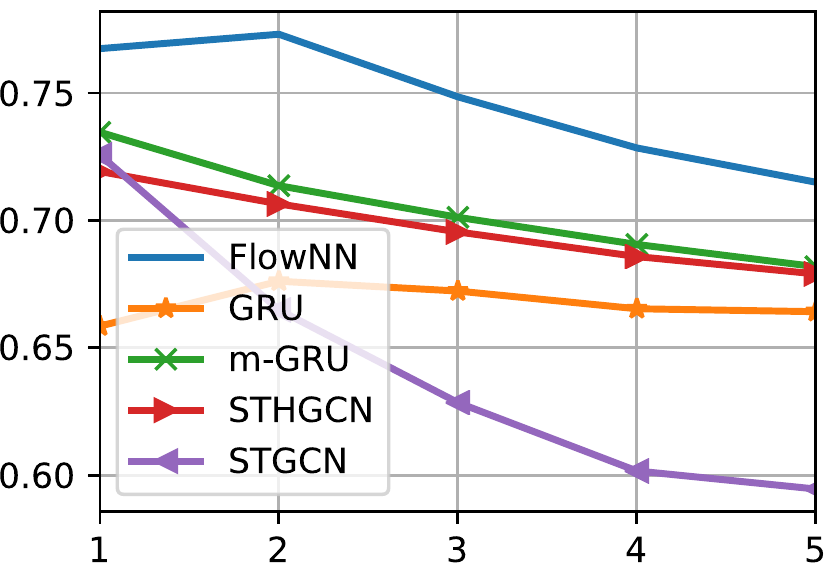}
\caption{Test Corr on \textit{WIDE}.}
\label{fig90}
\end{subfigure}%
\caption{Prediction test performances on unseen flows under different steps ahead.}
\label{figkstep}
\end{figure}
 \subsubsection{Multi-step-ahead rate predictions}
 Fig.~\ref{figkstep} shows the test performances of different models on multi-step-ahead prediction tasks. Following the practice of
second-granularity traffic forecasting in \cite{CNNdatacenter2018}, we aim to predict the average values across next multiple steps. That is, we predict the value of $\mathbb{E}[x^{t+1},\cdots, x^{t+\Delta}]$,
where $\Delta$ is the time length to predict. This information is crucial to many short-term networking planning and control tasks in communications networks. In this task, FlowNN is consistently best for different time predictions in the future. Note that for any $\Delta\geq2$, it is not always harder or easier to predict under the different degrees of traffic burstiness created by the average operation. Therefore, it is reasonable to observe that 2-step ahead prediction is better than the one-step ahead case in Fig.~\ref{figkstep}.

 \subsubsection{Efficacy of different recurrent iterations $N$} Table \ref{EfficacyIterN} presents the test results on unseen flows of dataset \textit{NUMFabric}. $N=2$ achieves the best results, but other values of $N$ do not degrade the  performance remarkably. This shows that extra iterations can further improve FlowNN. Although FlowNN is robust across different $N$'s, improperly setting $N$ will lead to very different computational complexity. Consequently, a smaller $N$ is preferred.
 \begin{table}[t]
\footnotesize
\centering
\caption{Efficacy of different iterations $N$ within FlowNN.}
\label{EfficacyIterN}
\begin{tabular}{|p{.1cm}|p{2.2cm}p{2.2cm}p{2.2cm}|}
\hline
          $N$            &MSE                            & RSE                    & Corr    \\  \hline
1             &   0.0274$\pm$0.0010 & 0.1711$\pm$0.0020 & 0.9853$\pm$0.0004  \\   \hline
2                &   \textbf{0.0261$\pm$0.0009} & \textbf{0.1672$\pm$0.0021} & \textbf{0.9860$\pm$0.0004}   \\  \hline
3           &    0.0273$\pm$0.0009 & 0.1710$\pm$0.0021 & 0.9854$\pm$0.0004    \\  \hline
4                &  0.0286$\pm$0.0010 & 0.1747$\pm$0.0022 & 0.9849$\pm$0.0004   \\   \hline
5             &    0.0280$\pm$0.0009 & 0.1731$\pm$0.0022 & 0.9849$\pm$0.0004     \\  \hline
\end{tabular}
\end{table}
 \subsubsection{Computational complexity}
The computational complexity of FlowNN depends on the length of routing paths. In the above experiments, we used several network topologies (fat tree, NSF, and GM50) with routing path lengths ranging
from 2 to 7. Evaluated under all these different topologies, the reported maximum computation time for FlowNN is about 0.3ms with a single Nvidia P40 24GB GPU. Such speed is enough to support every milisecond prediction task. Moreover, many networking tasks only require predicting the average of the next few steps in practice, which further alleviates the requirement on computation speed when deploying FlowNN. The computation time depends on both the model complexity and the computing devices. We leave the evaluation on larger topologies to the future.

 \section{Reproducibility}
  \subsubsection{Details of datasets} \textit{NUMFabric} is a synthetic dataset collected by the NS-3 based simulator\footnote{\url{https://bitbucket.org/knagaraj/numfabric/src/master/}} according to the configurations in \cite{numfabric2016}.
  The original simulator is configured with a 3-layer fat-tree network topology. While we used the same simulator by replacing the network topology as NSFNet topology and Germany 50 nodes topology from the site\footnote{\url{http://www.knowledgedefinednetworking.org/}}
  to generate the synthetic datasets of \textit{NSF} and \textit{GM50}, respectively. Instead of synthetic traffic patterns, dataset \textit{WIDE} is collected with the realistic packet traces from \url{https://mawi.wide.ad.jp/mawi/}.

  For all datasets, we collected the real-time sending rates of 50 random flows, as well as the aggregated sending rates along the routing path at a time interval of $1ms$ with a total running length of 30000$ms$. The sending rates are calculated by counting the packets received/transmitted every $ms$ from corresponding flows at each node. For the first 40
  flows, we split the time length of $30000ms$ into training, validation and test by ratio 6:2:2. Therefore, only historical data were exposed during training. The evaluation and test processes were both conducted with future data. Finally, we also
  tested the learnt model on the remaining 10 flows (unseen flows) that are never seen during training and evaluation.
 \subsubsection{Details of experiments}
For all baselines, we conduct a grid search over the hyper-parameters. The hidden size is chosen among \{16, 32, 64, 128, 256\} for all datasets. For all applied GRUs, the number of GRU layers is chosen among \{1, 2, 3, 4\}. For FlowNN, the number $N$ of iterations is chosen among \{1, 2, 3, 4, 5\}. We used the library from \url{https://www.dgl.ai/} for the implementations of GCN in STHGCN. For STGCN, we followed the implementations released in \url{https://github.com/VeritasYin/STGCN_IJCAI-18}.

The pseudocode implementation of FlowNN is described in Algorithm \ref{alg:algorithm}.

\begin{algorithm}[tb]
\caption{FlowNN pseudocode}
\label{alg:algorithm}
\textbf{Input}: $\{x_n^\tau\}_{\tau=t_0,\cdots,t-1, n=1,\cdots,L}$ \# \textit{note}: $x_{r,n}^\tau \in x_n^\tau$\\
\textbf{Output}: $\{\hat{h}_n^\tau\}_{\tau=t, n=1,\cdots,L}$ \quad\quad\#\textit{predicted representation}
\begin{algorithmic}[1] 
\State $h_n^\tau = f_{L_1}(x_n^\tau),\forall \tau=t_0,\cdots,t-1; n=1,\cdots,L$
\For{all $N$ recurrent iterations}
\State  \#\textit{PathAggregator}
\Statex         $\hat{h}_1^{\tau} = f_{L_2}(\{h_n^\tau\}_{n=1,\cdots,L}), \forall \tau=t_0,\cdots,t-1$
         \For{all neighboring pairs $\langle n,n+1\rangle_{n=1,\cdots,L-1}$}
         \Statex    $\langle T_1^i,T_2^i \rangle_{i=1,2,\cdots}=\mathrm{\textbf{\textit{split}}}(\{x_{r,n}^\tau-x_{r,n+1}^\tau\}_{\tau=t_0,\cdots,t-1})$
         \Statex   $\{\hat{h}_{n+1}^\tau\}_{\tau=t_1,\cdots,t}= \mathrm{\textbf{\textit{Induction}}}\big{(}\langle T_1^i,T_2^i \rangle_{i=1,2,\cdots},$
         \State       $\quad\quad\quad\quad\quad\quad\quad\quad\quad\ \ \{\hat{h}_1^\tau, h_{n+1}^\tau\}_{\tau=t_0,\cdots,t-1}\big{)}$
         \EndFor

\EndFor

\Algphase{Def $\mathrm{\textit{split}}\big{(}\big{)}$}
\textbf{Input}: $\{x_{r,n}^\tau-x_{r,n+1}^\tau\}_{\tau=t_0,\cdots,t-1} $
\State set correlation window index $i=0,T_1^i=T_2^i=[~]$
\For{all $\tau=t_0,\cdots,t-1$}
    \If{$x_{r,n}^\tau-x_{r,n+1}^\tau\geq 0$ }
    \State append $\tau$ to $T_1^i$
    \EndIf
    \If{$x_{r,n}^\tau-x_{r,n+1}^\tau < 0$}
    \State append $\tau$ to $T_i^2$
    \EndIf
    \If{$x_{r,n}^\tau-x_{r,n+1}^\tau < 0$ and $x_{r,n}^\tau-x_{r,n+1}^\tau \geq 0$}
    \State $i=i+1,T_1^i=T_2^i=[~]$
    \EndIf
\EndFor\\
\Return $\langle T_1^i,T_2^i \rangle_{i=1,2,\cdots}$

\Algphase{Def $\mathrm{\textit{Induction}}\big{(}\big{)}$}
\textbf{Input}: $\langle T_1^i,T_2^i \rangle_{i=1,2,\cdots}, \{\hat{h}_1^\tau, h_{n+1}^\tau\}_{\tau=t_0,\cdots,t-1}$
\State $\hat{h}_n^t = f_p(\hat{h}_1^t),\forall t\in\{T_1^i,T_2^i\}$
\For{each $i=1,2,\cdots$}
\State $\hat{h}^\tau_{n+1}=MLP\big{(}\{\hat{h}_{n}^{\tau}~||~h_{n+1}^{\tau}\}\big{)}, \forall \tau \in T_1^i$
\State $\hat{h}_{n+1}^\tau=Seq2Seq\big{(}h_{n+1}^{\tau-1}~ | ~\{\hat{h}_n^{t}\}_{t\in T_1^i}\big{)},\forall \tau\in T_2^i$
\EndFor\\
\Return $\{\hat{h}^\tau_{n+1}\}_{\tau=t_1,\cdots,t}$
\end{algorithmic}
\end{algorithm}

\section{Real-world Deployment}
\textbf{Path-wide data measurements:} As we explained in the definition of Packet Flow in Section \textbf{Preliminaries} of the main text, the features we analyzed in our model are $x_{f,n}^t\in \mathbb{R}$, which can be measured by telemetry monitors installed at each node. In the experiments, we collected the average packet transmission/receiving rates of each monitored flow at the nodes on its routing path to construct $x_{f,n}^t$, and predicted the corresponding transmission/receiving rates that a node will experience, and also the end-to-end transmission delay. Such prediction can be performed for each individual flow or aggregated flows that share the same routing path (including the source and destination nodes). There are many telemetry frameworks that support either per-flow or aggregated flow measurements, like INT \cite{INT2021} and Sketch \cite{sketch2020}.

\noindent\textbf{Real-world deployment:} In a real-world networking environment, the FlowNN model can be deployed at the end host/network interface controller (NIC) or a central controller. The real-time measurement can be collected by the northbound interface or ACK signals. Considering the deployment cost, the FlowNN model can be deployed to serve only selected flows that we really want to optimize, such as the networking flows from important clients. Other flows that only enjoy the best-effort services can be treated as background flows. This meets the commercial practice of network operators.

\bibliographystyle{named}
\bibliography{ijcai22}

\end{document}